\documentclass[preprint]{elsarticle}

\usepackage{lineno,hyperref}

\hypersetup{
    colorlinks=true,
    linkcolor=blue,
    filecolor=magenta,      
    urlcolor=cyan,
}

\modulolinenumbers[5]

\usepackage{graphicx}
\usepackage{amsmath}
\usepackage{hyperref} 

\begin{document}
\begin{frontmatter}

\title{A novel database of children's spontaneous facial expressions (LIRIS-CSE)}

\author[rvt,focal]{Rizwan Ahmed khan\corref{cor1}}
\author[rvt]{Crenn Arthur}
\author[rvt]{Alexandre Meyer}
\author[rvt]{Saida Bouakaz}

\cortext[cor1]{Corresponding author}
\address[rvt]{Faculty of IT, Barrett Hodgson University, Karachi, Pakistan.}
\address[focal]{LIRIS, Universit\'e Claude Bernard Lyon1, France.}

\begin{abstract}
Computing environment is moving towards human-centered designs instead of computer centered designs and human's tend to communicate wealth of information through affective states or expressions. Traditional Human Computer Interaction (HCI) based systems ignores bulk of information communicated through those affective states and just caters for user’s intentional input. Generally, for evaluating and benchmarking different facial expression analysis algorithms, standardized databases are needed to enable a meaningful comparison. In the absence of comparative tests on such standardized databases it is difficult to find relative strengths and weaknesses of different facial expression recognition algorithms. In this article we present a novel video database for Children's Spontaneous facial Expressions (LIRIS-CSE). Proposed video database contains six basic spontaneous facial expressions shown by 12 ethnically diverse children between the ages of 6 and 12 years with mean age of 7.3 years. To the best of our knowledge, this database is first of its kind as it records and shows spontaneous facial expressions of children. Previously there were few database of children expressions and all of them show posed or exaggerated expressions which are different from spontaneous or natural expressions. Thus, this database will be a milestone for human behavior researchers. This database will be a excellent resource for vision community for benchmarking and comparing results. In this article, we have also proposed framework for automatic expression recognition based on convolutional neural network (CNN) architecture with transfer learning approach. Proposed architecture achieved average classification accuracy of 75\% on our proposed database i.e. LIRIS-CSE.


\end{abstract}

\begin{keyword}
\texttt{Facial expressions database\sep spontaneous expressions\sep convolutional neural network\sep expression recognition\sep transfer learning.}
\end{keyword}

\end{frontmatter}

\section{Introduction}\label{sec:introduction}
Computing paradigm has shifted from computer-centered computing to human-centered computing \cite{N1}. This paradigm shift has created tremendous opportunity for computer vision research community to propose solution to existing problems and invent ingenious applications and products which were not though of before. One of the most important property of human-centered computing interfaces is the ability of machines to understand and react to social and affective or emotional signals \cite{N2,N10}.

Mostly humans express their emotion via facial channel, also known as facial expressions \cite{N10}. Humans are blessed with the amazing ability to recognize facial expression robustly in real-time but for machines it still is a difficult task to decode facial expressions. Variability in pose, illumination and the way people show expressions across cultures are some of the parameters that make this task more difficult \cite{KHAN2013_jPat}. 

Another problem that hinders the development of such system for real world applications is the lack of databases with natural displays of expressions \cite{112}. There are number of publicly available benchmark databases with posed displays of the six basic emotions \cite{53} i.e. happiness, anger, disgust, fear, surprise and sadness,  exist but there is no equivalent of this for spontaneous / natural basic emotions. While, it has been proved that spontaneous facial expressions differ substantially from posed expressions \cite{111}.

Another issue with most of publicly available databases is absence of children in recorded videos or images. Research community has put lot of efforts to built databases of emotional videos or images but almost all of them contain adult emotional faces \cite{91,88}. By excluding children's stimuli in publicly available databases vision research community not only restricted itself to application catering only adults but also produced limited study for the interpretation of expressions developmentally \cite{N5}.

\section{Publicly available databases of children's emotional stimuli}

To the best of our knowledge there are only three  publicly available databases that contains children emotional stimuli / images. They are:

\begin {enumerate}

\item The NIMH Child Emotional Faces Picture Set (NIMH-ChEFS) \cite{N6}
\item The Dartmouth Database of Children’s Faces \cite{N7}
\item The Child Affective Facial Expression {(CAFE)} \cite{N5}

\end {enumerate}

\begin{figure*}[!h]
\centering
\includegraphics[scale=0.23]{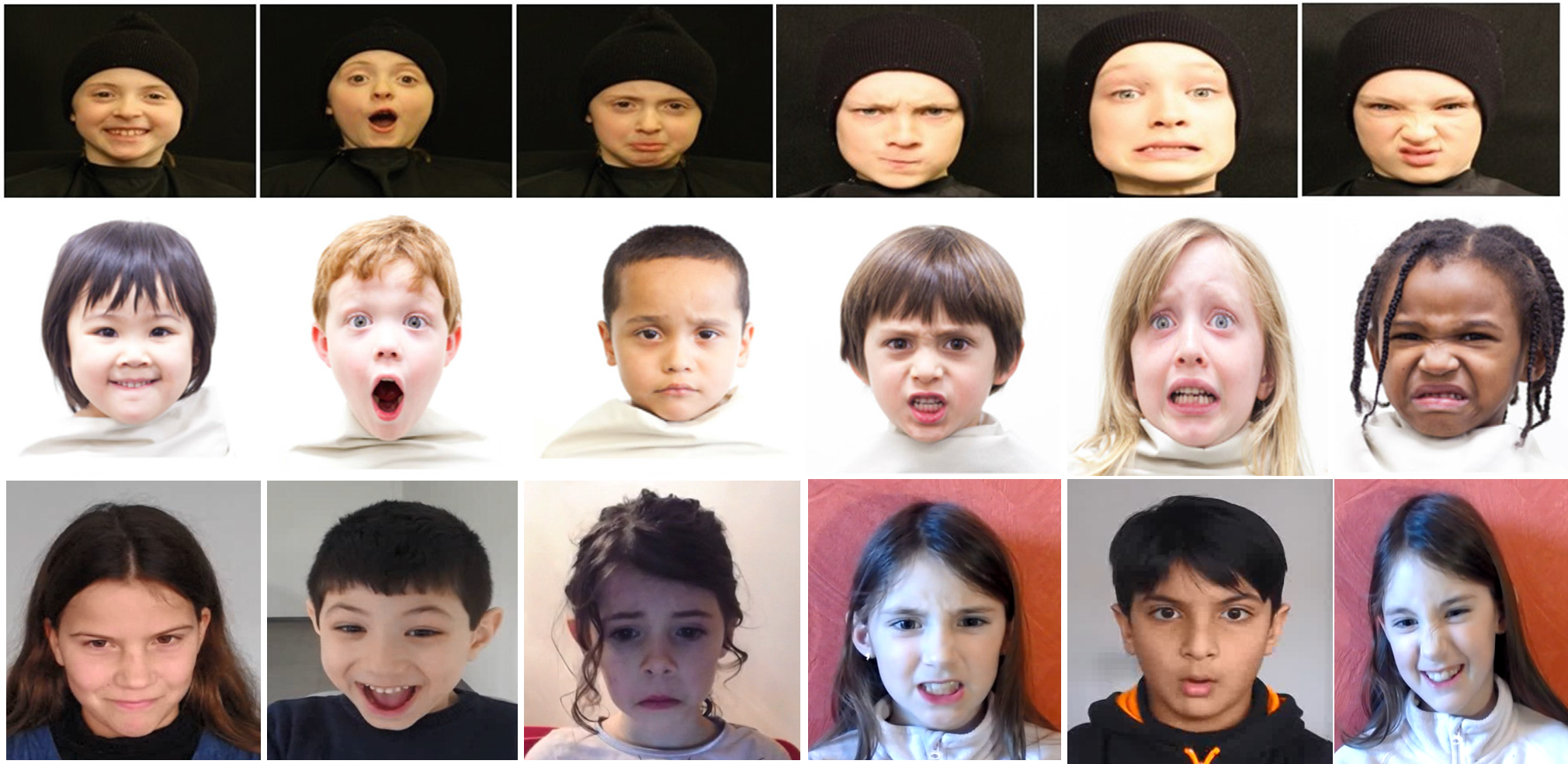}
\caption{ \textbf{Six universal expressions}: first row show example images from the Dartmouth database \cite{N7}. Second row show emotional images from the Child Affective Facial Expression {(CAFE)} \cite{N5}, while last row show emotional images from the movie clip of \textbf{our proposed database, LIRIS-CSE}. First column corresponds to expression of ``happiness'', second column corresponds to expression of ``surprise'', third column corresponds to expression of ``sadness'', fourth column corresponds to expression of ``anger'', fifth column corresponds to expression of ``fear'' and last column corresponds to expression of ``disgust''. Expressions in our proposed database are spontaneous and natural and can easily be differentiated from posed  / exaggerated expressions of the other two database. }\label{fig:DBcomp}
\end{figure*}

The NIMH Child Emotional Faces Picture Set (NIMH-ChEFS) \cite{N6} database has 482 emotional frames containing expressions of ``fear'', ``anger'', ``happy'' and ``sad'' with two gaze conditions: direct and averted gaze. Children that posed for this database were between 10 and 17 years of age.  The databases is validated by 20 adult raters.\\

The Dartmouth Database of children Faces \cite{N7} contains emotional images (six basic emotions) of 40 male and 40 female Caucasian children between the ages of 6 and 16 years. All facial images in the database were assessed by at least human 20 raters for facial expression identifiability and intensity. Expression of happy was most accurately identified while fear was least accurately identified by human raters. Human raters correctly classified 94.3\% of the happy faces while expression of fear was correctly identified in 49.08\% of the images, least identifiable by human raters. On average human raters correctly identified expression in 79.7\% of the images. Refer Figure \ref{fig:DBcomp} for examples images from the database.\\

The Child Affective Facial Expression (CAFE) database \cite{N5} is composed of 1192 emotional images (six basic emotions and neutral) of 2 to 8 years old children. Children that posed for this database were ethnically and racially diverse. Refer Figure \ref{fig:DBcomp} for examples frames from the database.

\subsection {Weaknesses of publicly available databases of children's emotional stimuli} \label{weakness}

Although above describe children expression databases are diverse in terms of pose, camera angles and illumination but have following drawbacks:

\begin {enumerate}

\item Above mentioned databases contain posed expressions and as mentioned before that spontaneous or natural facial expressions differ substantially from posed expressions as they exhibit real emotion whereas, posed expressions are fake and disguise inner feelings \cite{111, N11}.

\item All of these databases contains only static images / mug shots with expression at peak intensity. According to study conducted by psychologist Bassili \cite{N8} it was concluded that facial muscle motion/movement is fundamental to the recognition of facial expressions. He also concluded that human can robustly recognize expressions from video clip than by just looking at mug shot.

\item All of the above mentioned databases for children facial expressions present few hundred to maximum 1200 static frames. In current era where requirement of amount of data for learning a concept / computational model (deep learning \cite{120}) has increased exponentially, only few hundred static images are not enough. 

\end {enumerate}

\begin{figure*}[!h]
\centering
\includegraphics[scale=0.45]{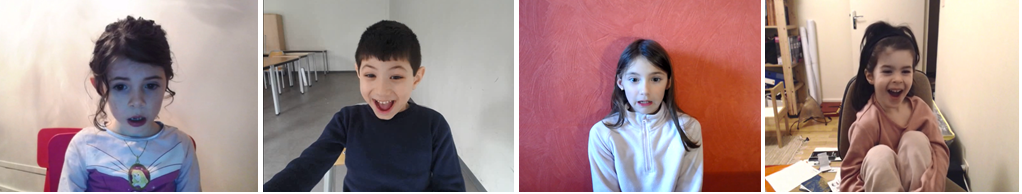}
\caption{\textbf{Example of variations in illumination condition and background}. Clips in column 1, 3 and 4 were recorded in home condition while image in column 2 was recorded in lab / classroom environment. }\label{fig:bg}
\end{figure*}

Generally, for evaluating and benchmarking different facial expression analysis algorithms, standardized databases are needed to enable a meaningful comparison. In the absence of comparative tests on such standardized databases it is difficult to find relative strengths and weaknesses of different facial expression recognition algorithms. Thus, it is utmost important to develop natural / spontaneous emotional database contains children movie clip / dynamic images. This will allow research community to built robust system for children's natural facial expression recognition. Thus, our \textit{contributions in this study are two-fold}: 

\begin {enumerate}

\item We are presenting a novel emotional database (LIRIS-CSE) that contains 208 movie clip / dynamic images of 12 ethnically diverse children showing spontaneous expressions in two environments, i.e. 1) lab / classroom environment 2) home environment (refer Figure \ref{fig:bg}). 

\item We have also proposed a framework for automatic facial expression recognition based on convolutional neural network (CNN) architecture with transfer learning approach. Proposed architecture achieved average classification accuracy of 75\% on our proposed database.

\end {enumerate}

\section{Novelty of proposed database (LIRIS-CSE)} \label{novel}

\begin{figure*}[!h]
\centering
\includegraphics[scale=0.45]{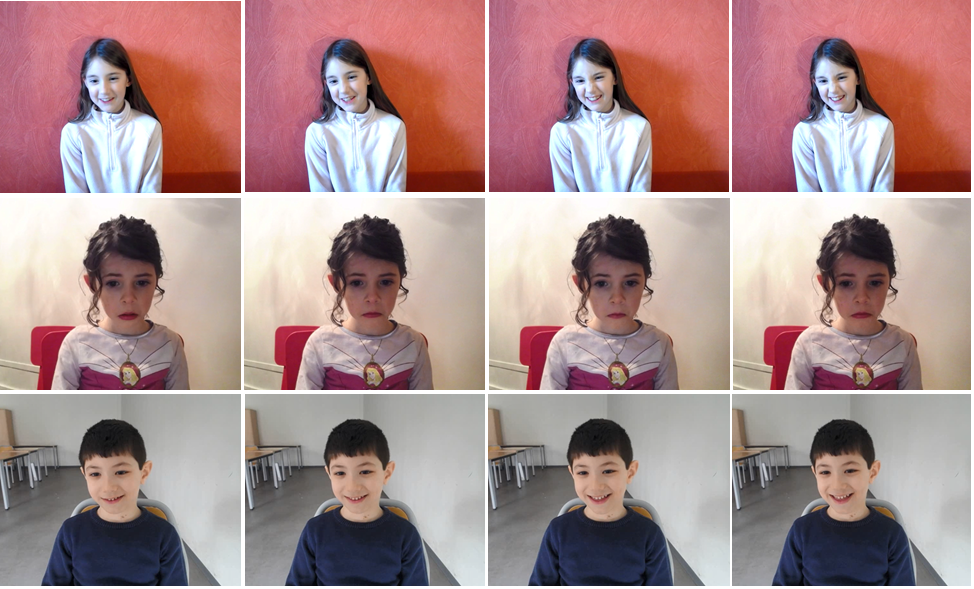}
\caption{\textbf{Example of expression transition}. First row shows example of expression of ``Disgust''. Second row shows expression of ``Sadness'', while third row corresponds to expression of ``Happiness''. }\label{fig:ExpTrans}
\end{figure*}

To overcome above mentioned drawbacks of databases of children's facial expression (refer Section \ref{weakness}), we are presenting a novel emotional database that contains movie clip / dynamic images of 12 ethnically diverse children. This unique database contains spontaneous / natural facial expression of children in diverse settings (refer Figure \ref{fig:bg} to see variations in recording scenarios) showing six universal or  prototypic emotional expressions (``happiness'', ``sadness'', ``anger'', ``surprise'', ``disgust'' and ``fear'') \cite{N15,48}. Children are recorded in constraint free environment (no restriction on head movement, no restriction on hands movement, free sitting setting, no restriction of any sort) while they watched specially built / selected stimuli. This constraint free environment allowed us to record spontaneous / natural expression of children as they occur. The database has been validated by 22 human raters. Details of recording parameters are presented in Table \ref{table:recParam}. In comparison with above mentioned databases for children facial expressions that have only few hundred images, our database (LIRIS-CSE) contains 26 thousand (26,000) frames of emotional data, refer Section \ref{vidSeg} for details.

The spontaneity of recorded expressions can easily be observed in Figure \ref{fig:DBcomp}. Expressions in our proposed database are spontaneous and natural and can easily be differentiated from posed  / exaggerated expressions of the other two databases. Figure \ref{fig:ExpTrans} shows facial muscle motion / transition for different spontaneous expressions. 


\subsection{Participants} \label{partici}
In total 12 (five male and seven female children) ethnically diverse children between the ages of 6 and 12 years with mean age of 7.3 years participated in our database recording session. 60 \% of recordings are done in classroom / lab environment, while 40\% of the clips in the database are recorded in home conditions. Recording children in two different environments has been done to have different background and illumination conditions in the recorded database. Refer Figure \ref{fig:bg} for example images with different backgrounds and illumination conditions.

\section{Database acquisition details} \label{DB-Acq}

First step for the creation of proposed spontaneous expressions database was the selection of visual stimuli that can induce emotions in children. Considering ethical reasons and young age of children we carefully selected stimuli and removed any stimuli that can have long term negative impact on the children. Due to these ethical reasons we did not include emotion inducer clips for the negative expression of ``anger'' and selected very few clips to induce emotion of ``fear'' and ``sadness''. The same has been practiced before by Valstar et. al \cite{N9}. Due to this very reason the proposed database contains more emotional clips of expressions of ``happiness'' and ``surprise''. Although there were no emotion inducer clips for the expression of ``anger'' but still database contains very few clips where children show expression of ``anger'' (refer Figure \ref{fig:DBcomp}) due to the fact that young children use expressions of ``disgust'' and ``anger'' interchangeably \cite{N12}.

\subsection{Emotion inducing stimuli} \label{EmotionalStimuli}
We either selected only animated cartoon / movies or small video clips of kids doing funny actions to stimuli list. The reasons for selecting videos to induce emotions in children are as follows:

\begin{enumerate}

\item All the selected videos for inducing emotions contains audio as well. Video stimuli along with audio gives immersive experience, thus is powerful emotion inducer \cite{N13}.  \\

\item Video stimuli provides more engaging experience then static images, restricting undesirable head movement. \\

\item Video stimuli can evoke emotions for a longer duration. This helped us in recording and spotting children facial expressions.
\end{enumerate}

List of stimuli selected as emotion inducers are presented in Table \ref{table:1}. Total running length of selected stimuli is 17 minutes and 35 seconds. One of the consideration for not selecting more stimuli is to prevent children's lose of interest or disengagement over time \cite{toAddinThesis6}.

\begin{table*}[!htb]
\centering

\begin{tabular}{ |p{1cm}|p{2.7cm}|p{1.5cm}|p{5.4cm}|p{1cm}|  }
 \hline
 \textbf{Sr.No:}	& \textbf{Induced Expression}	& \textbf{Source}	& \textbf{Clip Name}	&  \textbf{Time}\\
& & & & \\ \hline \hline

1	&Disgust	&YouTube	& Babies Eating Lemons for the First Time Compilation 2013	& 42 Sec \\ \hline
2	&Disgust	&YouTube	&On a Plane with Mr Bean (Kid puke)	&50 Sec \\ \hline
3	& Fear and surprise	& YouTube	& Ghoul Friend - A Mickey Mouse Cartoon - Disney Shows	& 50 Sec \\ \hline
4	& Fear	& YouTube	& Mickey Mouse - The Mad Doctor - 1933	& 57 Sec\\ \hline
5	& Fear \& surprise	& Film	&  “How To Train Your Dragon” (Monster dragon suddenly appears and kills small dragon)	& 121 Sec\\ \hline
6	& Fear	& Film	&  “How To Train Your Dragon” (Monster dragon throwing fire)	& 65 Sec\\ \hline
7	& Fear	& YouTube	& Les Trois Petits Cochons	& 104 Sec\\ \hline
8	& Happy	& YouTube	& Best Babies Laughing Video Compilation 2014 (three clips)	& 59 Sec\\ \hline
9	& Happy	& YouTube	& Tom And Jerry Cartoon Trap Happy 	& 81 Sec\\ \hline
10	& Happy, surprise \& fear	& YouTube	& Donald Duck- Lion Around 1950	& 40 Sec\\ \hline
11	& Happily surprised	& YouTube	& Bip Bip et Coyote - Tired and feathered	& 44 Sec\\ \hline
12	& Happily surprised	& YouTube	& Donald Duck - Happy Camping	& 53 Sec\\ \hline
13	& Sad	& YouTube	& Fox and the Hound - Sad scene	& 57 Sec\\ \hline
14	& Sad	& YouTube	& Crying Anime Crying Spree 3	& 14 Sec\\ \hline
15	& Sad	& YouTube	& Bulldog and Kitten Snuggling	& 29 Sec\\ \hline
16	& Surprise	& Film	& Ice Age- Scrat's Continental Crack-Up	& 32 Sec\\ \hline
17	& Surprise \& happy	& Film	& Ice Age (4-5) Movie CLIP - Ice Slide (2002)	& 111 Sec\\ \hline
18	& Happy	& YouTube	& bikes funny (3)	& 03 Sec\\ \hline
19	& Happy	& YouTube	& bikes funny	& 06 Sec\\ \hline
20	& Happy	& YouTube	 & The Pink Panther in 'Pink Blue Plate	& 37 Sec\\ \hline \hline
\multicolumn{5}{|r|}{\textbf{Total running length of stimuli = 17 minutes and 35 Seconds}} \\ \hline
\end{tabular}
\caption{Stimuli used to induce spontaneous expression}\label{table:1}
\end{table*}

\subsection{Recording setup}

\begin{figure}[!htb]
\centering
\includegraphics[scale=0.7]{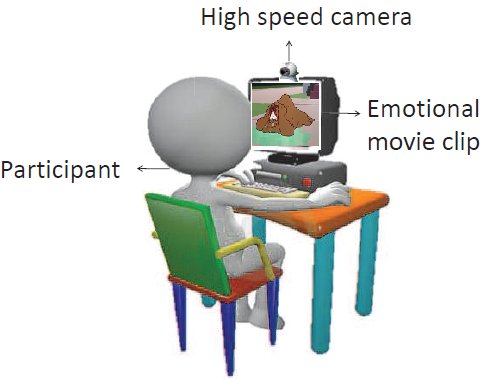}
\caption{ \textbf{Database recording setup}. Children watch emotions inducing stimuli while high speed webcam records their video focusing on face. Figure inspired by \cite{N13}.}\label{fig:SetUp}
\end{figure}

\begin{table}[h]
\centering
 \begin{tabular}{|l l l l|} 
 \hline

\textbf{Subject} &\textbf{Recording} 		&\textbf{FPS}					&\textbf{Video}					 \\
								 &\textbf{Environment}	&											&\textbf{Resolution}     \\
	\hline			
	\hline
	
S1-S7		&Classroom		&25						&800 * 600				\\
S8-S10	&Home					&25						&720 * 480				\\
S11-S12	&Home					&25						&1920 * 1080			\\

\hline
 \end{tabular}
\caption{Database videos recording parameters}\label{table:recParam}
\end{table}

Inspired by Li et al. \cite{N13}, we setup high speed webcam, mounted at the top of laptop with speaker output, at a distance of 50 cm. As explained above, The audio output enhanced the visual experience of a child, thus helping us induce emotions robustly. Recording setup is illustrated in Figure \ref{fig:SetUp}. As mentioned in Section \ref{partici}, children were recorded in two different environments i.e. classroom / lab environment and home environment. Details of recording parameters are presented in Table \ref{table:recParam}.

\subsection{Video segmentation}\label{vidSeg}


After recording video for each child we carefully examined the recoding and removed any unnecessary recorded part, usually at the beginning and at the end of video recording. As the video (with audio) stimuli that children watched was the combination of different emotional videos (refer Section \ref{EmotionalStimuli} for the details of visual stimuli), our recorded video contained whole spectrum of expressions in one single video. We then manually segmented one single video recording of each child into segments of small video chunks / clips such that each video clip show one pronounced expression. Refer Figure \ref{fig:ExpTrans} to see results after segmentation process. It can be observed from the referred figure that each small video clip contains neutral expression at the beginning, then shows onset of an expression, and finishes when expression is visible at its peak along with some frames after peak expression frame. Total number of small video clips, each containing specific expression, present in our database (LIRIS-CSE) are 208. Total running length of segmented clips in presented database (videos having children facial expressions) is seventeen minutes and twenty-four seconds. That makes total of around 26 thousand (26,000) frames of emotional data, considering recording is done at 25 frames / second, refer Table \ref{table:recParam}.  


\begin{figure*}[htb!]
\centering
\includegraphics[scale=0.466]{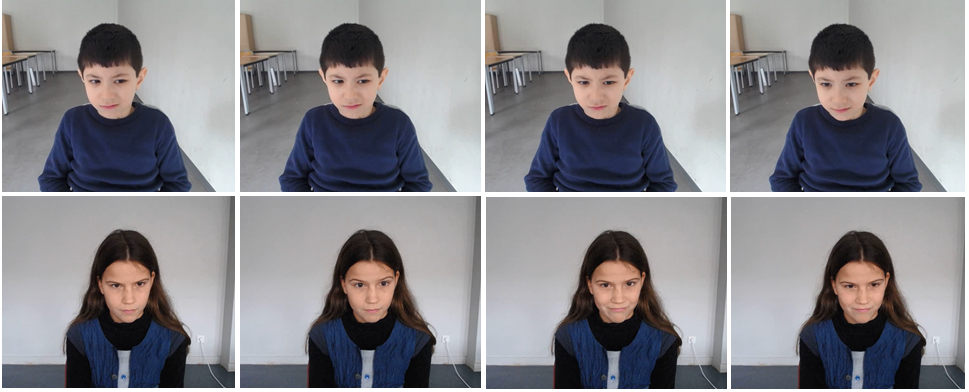}
\caption{\textbf{Blended Expressions}. Example images that show co-occurrence of more than one expressions in a single clip of the database. First row present frames from a clip that show occurrence of expressions of ``sadness'' and ``anger''. Similarly, second row shows co-occurrence of expressions of ``surprise'' and ``happiness''. }\label{fig:ExpConf}
\end{figure*}

There are seventeen (17) video clips present in this database that have two labels, for example ``happily surprised'', `` Fear surprise'' etc. This is due to the fact that for young children different expressions co-occur / blended expressions \cite{N14, N12} or a visual stimuli was so immersive that transition from one expression to another expression was not pronounced. Refer Figure \ref{fig:ExpConf} to see  example images from segmented video segments that show blended expressions.

\begin{figure}[htb!]
\centering
\includegraphics[scale=0.45]{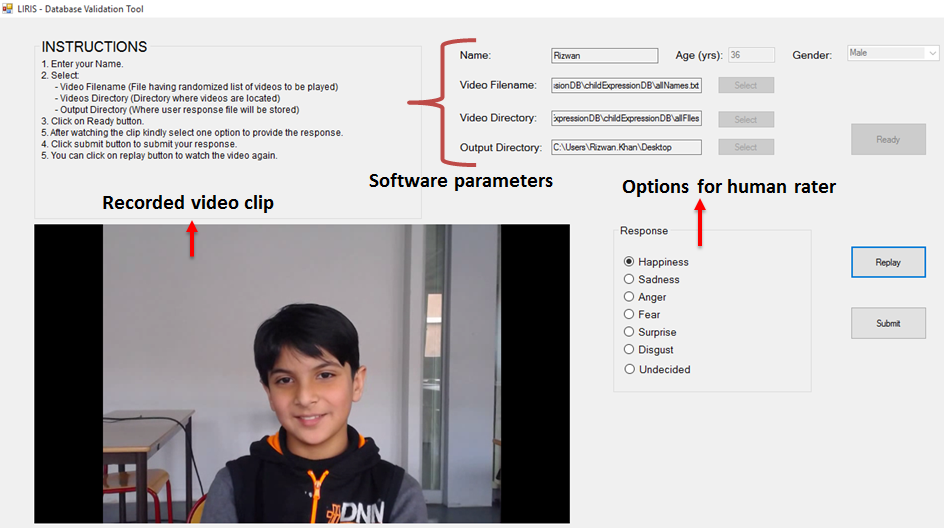}
\caption{\textbf{Validation tool.} Screen capture of validation tool used for collecting ratings / labels from human evaluators for each video clip. }\label{fig:VT}
\end{figure}

\subsection{Database validation}
The database has been validated by 22 human raters / evaluators between the ages of 18 and 40 years with mean age of 26.8 years. 50\% of database raters / evaluators were in the age bracket of $[$18 - 25$]$ years and rest of 50\% were in the age bracket of $[$26 - 40$]$ years. Human evaluators who were in the age bracket of $[$18 - 25$]$ years were university students and other group of evaluators were university faculty members. Human raters / evaluators were briefed about the experiment before they started validating the database. 

For database validation purpose we built software that played segmented video (in random order) and records human evaluator choice of  expression label. The screen capture of the software is presented in Figure \ref{fig:VT}. If required, evaluator can play video multiple times before recoding their choice for any specific video. 



In summary, instructions given to human raters / evaluators were following: 

\begin{enumerate}

\item Watch carefully each segmented video and select expression that is shown in the played segmented video. \\

\item If played video did not show any visible / pronounced expression, selected an option of  ``undecided''. Each video can be played multiple times without any upper bound on number of times video to be played. \\ 

\item Once expression label / response is submitted for a played segmented video then this label can not be edited.

\end{enumerate}

\begin{figure*}[ht]
\centering
\includegraphics[scale=0.44]{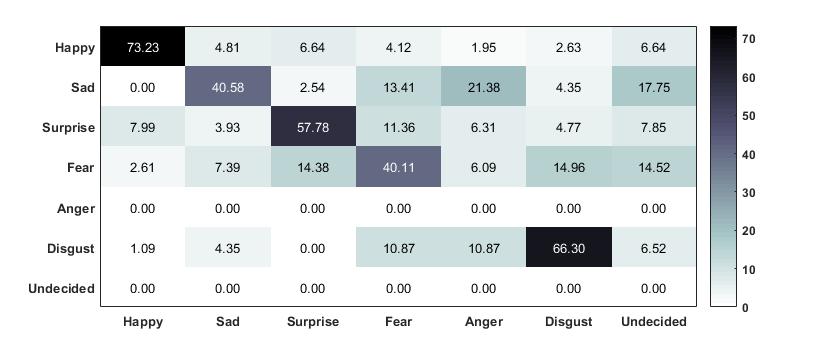}
\caption{\textbf{Confusion Marix}. Rows: induced intended expressions (average). Columns: expression label given by human raters (average). Diagonal represents agreement between induced intended expressions and expression label given by human raters, with darker colors representing greater agreement. }\label{fig:CM}
\end{figure*}

\subsubsection{Validation data analysis}

After validation data collection, we performed statistical analysis on the gathered data and calculated confusion matrix. Refer Figure \ref{fig:CM} to see calculated confusion matrix. Rows in the referred confusion matrix show induced intended expressions (average\%), while columns show expression label given by human raters / evaluators (average \%). Diagonal values represent agreement between induced intended expressions and expression label given by human evaluators, with darker colors representing greater agreement. 

As per calculated results expression of ``happy'' was most correctly spotted by evaluators, with average accuracy of 73.2\%. On the other hand expression of `` fear'' was least correctly identified by evaluators, with average accuracy of 40.1\%. These results are consistent with results from \cite{N7,N15}. We did not include expression of ``anger'' in analysis as there is only one movie clip with anger expression in the database.

Expression of ``fear'' which is least identified, is often perceptually mixed with expressions of ``surprise'' and ``disgust''. As mentioned above, this is due to the fact that for young children different expressions co-occur (blended expressions) \cite{N14, N12} or a visual stimuli was so immersive that transition from one expression to another expression was not pronounced. Refer Figure \ref{fig:ExpConf} to see  example images from segmented video segments that show blended expressions. 

Overall average accuracy of human evaluators / raters is 55.3\%. As per study published by Matsumoto et al. \cite{N16} human's usually can spot expressions correctly 50\% of the time and the easiest expression for human's to identify are ``happy'' and ``surprise''. These results conforms well with the results that we obtained from human evaluators as expression of ``happy'' was most correctly identified while average accuracy of human evaluators raters is also around 50\% (55.3\% to be exact).

\section{Database availability}
The novel database of Children's Spontaneous Expressions (LIRIS-CSE) is available for research purposes only. It can be downloaded by researcher / lab after signing End User License Agreement (EULA). Website to download LIRIS-CSE database is:
\href{https://childrenfacialexpression.projet.liris.cnrs.fr/} {https://childrenfacialexpression.projet.liris.cnrs.fr/}.

\section {Automatic recognition of affect, a transfer learning based approach} \label{ML}

In order to provide benchmark machine learning / automatic classification of expression results on our database (LIRIS-CSE), we have done experiment based on transfer learning paradigm. Usually machine learning algorithms make prediction on data that is similar to what algorithm is trained on; training and test data are drawn from same distribution. On the contrary transfer learning allows distributions used in training and testing to be different \cite{tLearn}. We used transfer learning approach in our experiment due to following facts: 

\begin{enumerate}

\item We wanted to benefit from deep learning model that has achieved high accuracy on recognition tasks that take image as input i.e. ImageNet Large-Scale Visual Recognition Challenge (ILSVRC) \cite{ILSVRC15},  and is available for research purposes. 

\item It requires large database to train deep learning model from the scratch \cite{DL}. In a given scenario, we can not train deep learning model from the very beginning. 

\end{enumerate}

\subsection{Convolutional Neural Network (CNN)}\label{CNNsec}
Researchers have been successful in developing models that can recognize affect robustly \cite{KHAN2013_jPat,Khan_ISVC,Khan2018}. Recently, most of successful models are based on deep learning approach \cite{117,118,119}, specifically on Convolutional Neural Network (CNN) architecture \cite{120}. CNNs are class of deep, feed forward neural networks that have shown robust results for applications involving visual input i.e image / object recognition \cite{125}, face expression analysis \cite{119}, semantic scene analysis / semantic segmentation \cite{125,127}, gender classification \cite{128} etc.


The architecture of CNN was first proposed by LeCun \cite{120}. It is a multi-stage or multi-layer architecture. This essentially means there are multiple stages in CNN for feature extraction. Every stage in the network has an input and output which is composed of arrays known as feature maps. Every output feature map consists of patterns or features extracted on locations of the input feature map. Every stage is made up of layers after which classification takes place \cite{corr1803, zeilerconv,123}. Generally, these layers are:

\begin{enumerate}

\item Convolution layer: This layer makes use of filters, which are convolved with the image, producing activation or feature maps. 
\item Feature Pooling layer: This layer is inserted to reduce the size of the image representation, to make the computation efficient. The number of parameters is also reduced which in turn controls over-fitting. 
\item Classification layer: This is the fully connected layer. This layer computes the probability / score learned classes from the extracted features from convolution layer in the preceding steps.

\end{enumerate}

\begin{figure*}
\centering
\includegraphics[scale=0.45]{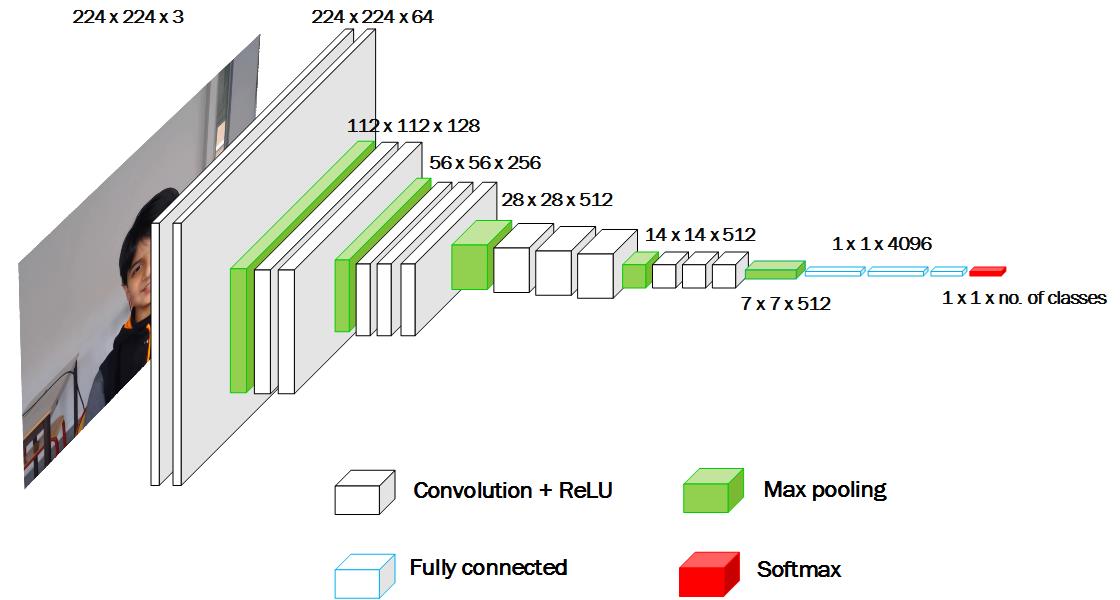}
\caption{An illustration of VGG16 architecture \cite{150}.} \label{VGG16}
\end{figure*}

\subsection{VGGNet architecture and transfer learning}\label{VGGsec}

Since 2012, deep Convolution Networks (ConvNets) have become a focus of computer vision scientists. Various architectures were proposed to achieve higher accuracy for a given tasks. For example,  best submissions to ImageNet Large-Scale Visual Recognition Challenge (ILSVRC) \cite{ILSVRC15}, \cite{151,152} proposed to use a smaller receptive window size / smaller filter size and smaller stride in the first convolutional layer. Generally, ImageNet Large-Scale Visual Recognition Challenge (ILSVRC) has served as a platform for advancements in deep visual recognition architectures.

The best proposed ConvNets architectures for ILSVRC 2014 competition were GoogleNet (a.k.a. Inception V1) from Google \cite{GoogleNet} and VGGNet by Simonyan and Zisserman \cite{150}. GoogleNet contains 1 x 1 Convolution at the middle of the network and global average pooling was used at the end of the network instead of using fully connected layers, refer Section \ref{CNNsec} for discussion on different types of layers. VGGNet consists of 16 convolutional layers (VGG16). It is one of the most appealing framework because of its uniform architecture, refer Figure \ref{VGG16}. It's pre-trained model is freely available for research purpose, thus making a good choice for transfer learning.

VGG16 architecture (refer Figure \ref{VGG16}) takes image of 224 x 224 with the receptive field size of 3 x 3. The convolution stride is 1 pixel and padding is 1 (for receptive field of 3 x 3). There are two fully connected layers with 4096 units each, the last layer is a softmax classification layer with $x$ units (representing $x$ classes / $x$ classes to recognize) and the activation function is the rectified linear unit (ReLU). The only downside of VGG16 architecture is its huge number of trainable parameters. VGG16 consists of 138 million parameters.

\subsection {Experimental framework and results} \label{ML}

As discussed earlier, CNN requires large database to learn concept \cite{129,DL}, making it impractical for different applications. This bottleneck is usually avoided using transfer learning technique \cite{121}. Transfer learning is a machine learning approach that focuses on ability to apply relevant knowledge from previous learning experiences to a different but related problem. We have used transfer learning approach to built framework for expression recognition using our  proposed database (LIRIS-CSE) as the size of our database is not sufficiently large to robustly train all layers of CNN from the very beginning. We used pre-trained VGG model (VGG16, a 16 layered architecture) \cite{150}, which is a deep convolutional network  trained for object recognition \cite{125}. It is developed and trained by Oxford University's Visual Geometry Group (VGG) and shown to achieve robust performance on the ImageNet dataset \cite{126}for object recognition. Refer Section \ref{VGGsec} for discussion on VGG16.

\begin{figure*}
\centering
\includegraphics[scale=0.43]{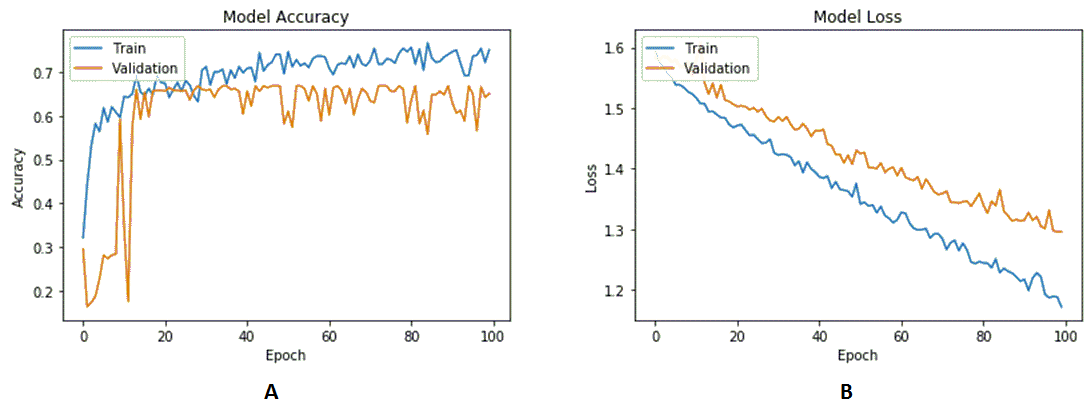}
\caption{CNN model learning: (A) Training accuracy vs Validation accuracy (B) Training loss vs Validation loss.} \label{CNN-accuracy}
\end{figure*}

We replaced last fully connected layer of VGG16 pre-trained model with dense layer having five outputs. This makes 5005 trainable parameters. Number of output of last dense layer corresponds to number of classes to be recognized, in our experiment we learned concept of five classes i.e. five expression to be recognized (out of six universal expression, expression of ``anger'' was not included in this experiment as there are few (one) clip(s) for ``anger'', for explanation see Section \ref{DB-Acq}). We trained last dense layer with images (frames from videos) from our proposed database using softmax activation function and ADAM optimizer \cite{130}. 

Our proposed database consists of video but for this experiment we extracted frames from videos and fed them to above described ConvNet architecture. We used 80\% of frames for training and 10\% of frames for validation process. With above mentioned parameters, proposed CNN achieved average expression accuracy of 75\% on our proposed database (five expressions). Model accuracy and loss curves are shown in Figure \ref{CNN-accuracy}.

\section{Conclusion}
In this article we presented novel database for Children's Spontaneous Expressions (LIRIS-CSE). The database contains six universal spontaneous expression shown by 12 ethnically diverse children between the ages of 6 and 12 years with mean age of 7.3 years. There were five male and seven female children. 60\% of recordings were done in classroom / lab environment and 40\% of the clips in the database were recorded in home conditions.

The LIRIS-CSE database contains 208 small video clips (on average each clip is 5 seconds long), with each clip containing one specific expression. Clips have neutral expression / face at the beginning of clip, then it show onset of an expression, and finishes when expression is visible at its peak along with some frames after peak expression frame. The database has been validated by 22 human raters / evaluators between the ages of 18 and 40 years. 

To the best of our knowledge, this database is first of its kind as it records and shows six (rather five as expression of ``anger'' is spotted/recorded only once) universal spontaneous expressions of children. Previously there were few  image databases of children expressions and all of them show posed or exaggerated expressions which are different from spontaneous or natural expressions. Thus, this database will be a milestone for human behavior researchers. This database will be a excellent resource for vision community for benchmarking and comparing results. 

For benchmarking automatic recognition of expression we have also provided result using Convolutional Neural Network (CNN) architecture with transfer learning approach. Proposed approach obtained average expression accuracy of 75\% on our proposed database, LIRIS-CSE (five expressions).

\section*{References}


\end{document}